\documentclass[letterpaper]{article} % DO NOT CHANGE THIS
\usepackage{aaai2027} % DO NOT CHANGE THIS
\usepackage[hyphens]{url} % DO NOT CHANGE THIS
\usepackage{graphicx} % DO NOT CHANGE THIS
\usepackage{natbib} % DO NOT CHANGE THIS
\usepackage{caption} % DO NOT CHANGE THIS
\usepackage{array}
\usepackage{booktabs}
\nocopyright

\usepackage{booktabs}
\usepackage{amsmath}
\usepackage{amsfonts}
\usepackage{nicefrac}
\usepackage{microtype}
\usepackage{xcolor}
\usepackage{multirow}
\usepackage{makecell}
\usepackage{array}
\usepackage{tabularx}
\usepackage{adjustbox}
\usepackage{soul}

\title{The Parts Are Greater Than the Sum: Automated Task Sequencing for Efficient Training of Multi-Policy LLMs}

\author{ Jiajia Tang,\textsuperscript{1}
    Sizhe Yuen,\textsuperscript{1}
    Francisco Gomez Medina,\textsuperscript{1}
    Yali Du, \textsuperscript{1,3}
    Adam Sobey\textsuperscript{1,2}}
\affiliations{
    \textsuperscript{1}The Alan Turing Institute, London, UK\\
    \textsuperscript{2}University of Southampton, Southampton, UK\\
    \textsuperscript{3}King’s College London, London, UK\\

}

\begin{document}

\maketitle

\begin{abstract}
Parameter-Efficient Fine-Tuning (PEFT) commonly adapts large language models using a single shared Low-Rank Adapter (LoRA). This shared optimization space often suffers from interference when adapting heterogeneous task sequences, leading to poor transfer and catastrophic forgetting. Existing approaches mainly improve adapter expressiveness by increasing parameter capacity or composing multiple adapters, yet they still rely on a shared optimization path. In this paper, we propose an optimization-path organization framework for parameter-efficient fine-tuning of large language models, implemented as an automatic multi-policy PEFT architecture. Specifically, optimization-compatible adaptation paths are automatically organized through  task grouping and task sequencing under a fixed parameter budget. The organized optimization paths are implemented as independent Quantized Low-Rank Adapters (QLoRA), enabling heterogeneous tasks to be optimized in decoupled adaptation spaces while preserving positive transfer among compatible tasks. Experiments on the TRACE benchmark demonstrate that performance consistently improves from conventional single-policy PEFT to multi-policy PEFT, with the proposed automatic multi-policy framework achieving the best performance of 44.78 under the same trainable capacity. This suggests that optimization-path organization is more effective than simply increasing adapter capacity for heterogeneous parameter-efficient fine-tuning.

\end{abstract}

\section{Introduction}

Parameter-Efficient Fine-Tuning (PEFT) \citep{Ding2023-peft} has become the dominant paradigm for adapting large language models (LLMs) to downstream tasks. 
%Compared to full-parameter fine-tuning, PEFT updates only a small subset of trainable parameters while keeping the pretrained backbone frozen, significantly reducing memory consumption and storage overhead. 
Representative methods such as Low-Rank Adaptation (LoRA) and Quantized Low-Rank Adaptation (QLoRA) demonstrate that lightweight adaptation can achieve competitive downstream performance while maintaining high efficiency \cite{houlsby2019,hu2021,dettmers2023qlora}.
Despite their architectural differences, most existing PEFT methods rely on a common assumption: a single shared low-rank adapter is expected to accommodate all target tasks \citep{wang2023trace}. This assumption is reasonable when tasks have similar characteristics, allowing the shared adaptation space to promote positive transfer. However, under heterogeneous task sequences during parameter-efficient fine-tuning, tasks often differ substantially in domain, supervision signals, reasoning patterns, capability requirements, and output formats. Such heterogeneity frequently leads to inefficient learning \cite{michieli2024hop}. Consequently, forcing all tasks to update the same low-rank adaptation space introduces optimization interference, resulting in negative transfer and catastrophic forgetting. A few approaches have therefore focused on assigning a single LoRA to each task \cite{yang2025mtlloralowrankadaptationmultitask, zhang-etal-2025-mixture} but these approaches will lose efficiency from lack of transfer learning and increase in size for large numbers of tasks. 
%This limitation becomes particularly evident in heterogeneous continual learning benchmarks such as TRACE \citep{wang2023trace}, which combines diverse tasks including domain-specific question answering, multilingual generation, code completion, and mathematical reasoning. Standard PEFT methods experience substantial performance degradation when adapting these heterogeneous task streams sequentially. This observation raises a fundamental question for efficient continual adaptation of LLMs: is the performance bottleneck caused by insufficient adapter capacity, or by organizing all heterogeneous tasks into a single shared optimization path?

Existing studies improve adapter expressiveness through higher-rank adapters, richer parameterizations, or multi-adapter architectures such as adapter fusion, hypernetwork-based generation, and modular composition.\citep{pfeiffer2021adapterfusion,pfeiffer2020madx,mahabadi2021hyperformer,mahabadi2021compacter}. Although these approaches improve representation capability, they largely preserve a shared optimization mechanism, where heterogeneous tasks remain coupled through either a common parameter space or a shared adapter generation process. Consequently, they alleviate optimization interference without explicitly organizing heterogeneous optimization paths. The key challenge is not how many trainable parameters are used, but how heterogeneous tasks are organized before training. That is, the task grouping and task sequence should be taken into account, which is essential for constructing optimization-compatible adaptation paths under a fixed parameter budget. 
Unlike conventional multi-task learning, where tasks are jointly optimized, we consider parameter-efficient fine-tuning over heterogeneous task sequences, in which tasks are learned sequentially.

To address this challenge, we propose an optimization-path organization framework for parameter-efficient fine-tuning of large language models, implemented using a multi-policy PEFT architecture. We take our inspiration from multi-policy approaches in Reinforcement Learning which improve the efficiency of learning  \citep{bossens2024lifetimepolicyreuse,birkbeck2025chirpschangeinducedregretproxy}. The framework automatically organizes optimization-compatible adaptation paths through two consecutive stages. First, heterogeneous tasks are grouped according to optimization compatibility, reducing gradient conflicts while preserving capability similarity. 
Each task group is assigned an independent optimization path, implemented as a separate policy for parameter-efficient fine-tuning. Second, a task sequence is automatically constructed within each policy by organizing the learning sequence to minimize transition costs while reducing catastrophic forgetting. Consequently, task grouping and task sequencing encourage positive transfer among compatible tasks and produce smoother optimization trajectories throughout sequential fine-tuning. Unlike existing modular PEFT methods, which mainly improve adapter architectures or parameter composition, our framework explicitly organizes optimization paths before adaptation begins. Multiple independent QLoRA adapters serve as the implementation mechanism of the organized optimization paths, reducing optimization interference without introducing additional routing modules, fusion layers, or hypernetworks.

%Our contributions are summarized as follows:

%\begin{itemize}
%    \item We identify that the performance limitation of PEFT originates not only from limited adapter capacity but also from the organization of the shared optimization path, which introduces optimization interference among heterogeneous tasks.
    
 %   \item 
%    We propose an optimization-path organization framework for heterogeneous continual adaptation. The framework is implemented using a multi-policy PEFT architecture, where optimization-compatible adaptation paths are automatically constructed through task grouping and continual task sequencing.
   % We propose an automatic multi-policy PEFT framework that jointly optimizes task grouping and learning sequence to construct optimization-compatible adaptation policies under a fixed parameter budget. The framework automatically groups compatible tasks into independent QLoRA adapters and further learns an effective continual curriculum within each adapter.
%\end{itemize}

\section{Related Work}

\subsection{Parameter-Efficient Fine-Tuning}

Parameter-efficient fine-tuning (PEFT) aims to adapt large pretrained models by training only a lightweight set of additional parameters while keeping the backbone model largely frozen. Early adapter-based methods insert small trainable modules into Transformer layers and demonstrate that strong transfer performance can be achieved with very limited parameter overhead \citep{houlsby2019}. Subsequently, LoRA formulates parameter updates as low-rank decompositions, and QLoRA further combines low-rank adaptation with 4-bit quantization, making efficient adaptation feasible for much larger LLMs without sacrificing downstream performance \citep{hu2021,dettmers2023qlora}. Prompt-based PEFT methods, including Prefix-Tuning, Prompt Tuning, and P-Tuning v2, investigate lightweight adaptation through trainable prompts rather than trainable adapters \citep{li2021prefix,lester2021prompt,liu2022ptuningv2}. Other PEFT variants, such as IA$^3$, AdaLoRA, and UniPELT, further improve adaptation flexibility through multiplicative rescaling, adaptive rank allocation, or unified combinations of PEFT modules \citep{liu2022ia3,zhang2023adalora,mahabadi2022unipelt}. More recently, comprehensive surveys have summarized the rapid development of PEFT techniques, covering adapter tuning, prompt tuning, low-rank adaptation, and related lightweight optimization strategies for large language models \citep{han2024peftsurvey,wang2024peftsurvey}.

%Although these methods differ substantially in implementation, most of them assume that all target tasks can be optimized through a shared adaptation path. Such an assumption is effective when task distributions are relatively homogeneous. However, under heterogeneous task sequences during parameter-efficient fine-tuning, tasks often exhibit distinct optimization characteristics, capability requirements, and output formats, making a shared optimization space susceptible to optimization interference. In contrast to existing PEFT methods that mainly improve adapter expressiveness, our work focuses on automatically organizing optimization paths through joint task grouping and task sequencing before adaptation begins.
Although these methods differ substantially in implementation, most assume that all target tasks are optimized through a shared adaptation path. 
This shared optimization space works well for relatively homogeneous tasks, but becomes susceptible to optimization interference under heterogeneous task sequences. 
Instead of improving adapter expressiveness, our work automatically organizes optimization paths through joint task grouping and task sequencing before adaptation begins.

\subsection{Multi-Adapter PEFT}

Multiple adapters or richer composition mechanisms are a common approach to improving performance. AdapterFusion combines multiple pretrained task adapters through attention-based fusion during inference \citep{pfeiffer2021adapterfusion}. MAD-X composes language adapters and task adapters to support modular multilingual transfer \citep{pfeiffer2020madx}. HyperFormer generates task-aware adapters through shared hypernetworks, while Compacter improves adapter efficiency through structured parameter sharing \citep{mahabadi2021hyperformer,mahabadi2021compacter}. More recently, AdapterSoup explores parameter aggregation across multiple adapters to improve generalization \citep{chronopoulou2023adaptersoup}. Modular and expert-style PEFT architectures have also attracted increasing attention. MixLoRA introduces a LoRA-based mixture-of-experts architecture for efficient multi-task adaptation \citep{li2024mixlora}, while SLIM proposes a soft LoRA mixture mechanism that improves adaptation performance while reducing forgetting during continual fine-tuning \citep{han2025slim}. These studies demonstrate that adapter organization and modular design can substantially influence downstream adaptation performance.

Optimization interference has also been extensively studied in continual learning. Gradient-based methods, including PCGrad and Gradient Vaccine, alleviate optimization conflicts by modifying or regularizing task gradients during optimization \citep{yu2020pcgrad,wang2021gradient}. Parameter-isolation approaches, such as Progressive Networks and PackNet, reduce interference by assigning separate subnetworks or parameter subsets to different tasks \citep{rusu2016progressive,mallya2018packnet}. More recently, MoCL proposes a modular continual learning framework for language models, showing that structured parameter organization improves knowledge retention while maintaining transfer across tasks \citep{wang2024mocl}. Collectively, these studies suggest that optimization behavior is strongly influenced by how trainable parameters are organized.

%Unlike existing methods that primarily improve adapter expressiveness, manipulate gradients, or isolate parameters after optimization begins, our work addresses a complementary problem: how heterogeneous tasks should be organized before optimization starts. Rather than introducing more expressive adapters or additional routing mechanisms, our framework automatically organizes optimization paths by assigning compatible tasks to independent QLoRA adapters and constructing an effective task sequence within each adapter.
Unlike existing methods that improve adapter expressiveness, manipulate gradients, or isolate parameters, our work addresses the complementary problem of organizing heterogeneous tasks before optimization. Our framework automatically organizes optimization paths by assigning compatible tasks to independent QLoRA adapters and constructing an effective task sequence within each adapter.

\subsection{Sequential Fine-Tuning for Large Language Models}

Sequential fine-tuning has been studied in several closely related continual-learning settings, including continual instruction tuning, sequential domain adaptation, and heterogeneous task streams. TRACE provides a representative benchmark for studying continual learning across eight highly heterogeneous language tasks \citep{wang2023trace}. More recent benchmarks further extend this direction. For example, CoIN investigates continual instruction tuning for multimodal large language models and highlights the challenge of maintaining performance across sequential instruction streams \citep{chen2024coin}. InsCL proposes a data-efficient continual learning framework for instruction tuning and demonstrates that catastrophic forgetting remains a major challenge even under parameter-efficient adaptation \citep{wang2024inscl}. Furthermore, recent surveys provide comprehensive reviews of continual learning for large language models and parameter-efficient continual fine-tuning, covering continual pre-training, instruction tuning, domain adaptation, and task-specific continual learning \citep{shi2025clllms,coleman2025pecftsurvey}.

%Most existing continual learning studies primarily investigate how to reduce forgetting during sequential adaptation. In contrast, our work studies how heterogeneous tasks should be organized before parameter-efficient fine-tuning begins. The proposed framework therefore provides a complementary perspective based on optimization-path organization for sequential fine-tuning, rather than introducing a new continual learning mechanism.
Most continual learning studies focus on reducing forgetting during sequential adaptation. In contrast, our work studies how heterogeneous tasks should be organized before parameter-efficient fine-tuning. Rather than introducing a new continual learning mechanism, our framework provides a complementary perspective based on optimization-path organization for sequential fine-tuning.

\section{Automated Task Sequencing for Fine-Tuning of Multi-Policy LLMs}

\subsection{Problem Formulation}

We consider parameter-efficient fine-tuning of a pretrained large language model $f_{\theta}$, where the backbone parameters $\theta$ are frozen and only a small set of adaptation parameters are trained.
We focus on a pre-deployment setting in which the target task collection is available before fine-tuning begins.
Let $\mathcal{T} = \{T_1, T_2, \ldots, T_N\}$ denote a heterogeneous sequential task collection, where each task $T_i$ is associated with a data distribution $\mathcal{D}_i$ and a task-specific loss $\ell_i$. In standard LoRA-style PEFT, all tasks share one low-rank update,
\begin{equation}
\Delta \theta = AB,
\quad
A \in \mathbb{R}^{d \times r},
\quad
B \in \mathbb{R}^{r \times d},
\quad
r \ll d.
\end{equation}
The corresponding shared-adapter objective can be written as,
\begin{equation}
\mathcal{L}_{\mathrm{shared}}
=
\sum_{i=1}^{N}
\mathbb{E}_{(x,y)\sim \mathcal{D}_i}
\left[
\ell_i
\left(
f_{\theta+\Delta\theta}(x), y
\right)
\right].
\end{equation}

%This formulation requires all heterogeneous tasks to be optimized through the same low-rank adaptation space. When tasks are closely related, such sharing can encourage positive transfer. However, heterogeneous tasks may differ substantially in domain, capability requirement, and output format, resulting in incompatible optimization directions. 
Let $g_i = \nabla_{\Delta\theta}\ell_i$ denote the gradient induced by task $T_i$. If two tasks produce weakly aligned or conflicting gradients, their updates compete within the same trainable subspace. The shared adapter is consequently forced to follow a compromised optimization trajectory, which may be suboptimal for multiple tasks and leads to weak transfer or catastrophic forgetting. The bottleneck is therefore not merely insufficient adapter capacity. Even when the adapter rank is increased, all heterogeneous tasks remain constrained to co-update the same trainable space and follow the same coupled optimization path. Rather than simply enlarging this shared space, we reorganize the optimization process. 
%Our goal is to construct multiple policies that improve the \emph{optimization-compatibility of each},  separating strongly conflicting tasks while preserving useful transfer among compatible ones.

Our goal is to automatically organize optimization paths for heterogeneous task sequences during parameter-efficient fine-tuning by separating strongly conflicting tasks while preserving useful transfer among compatible ones. These organized optimization paths are implemented as multiple independent policies.

\begin{figure*}[tp]
    \centering
    \includegraphics[width=0.9\linewidth]{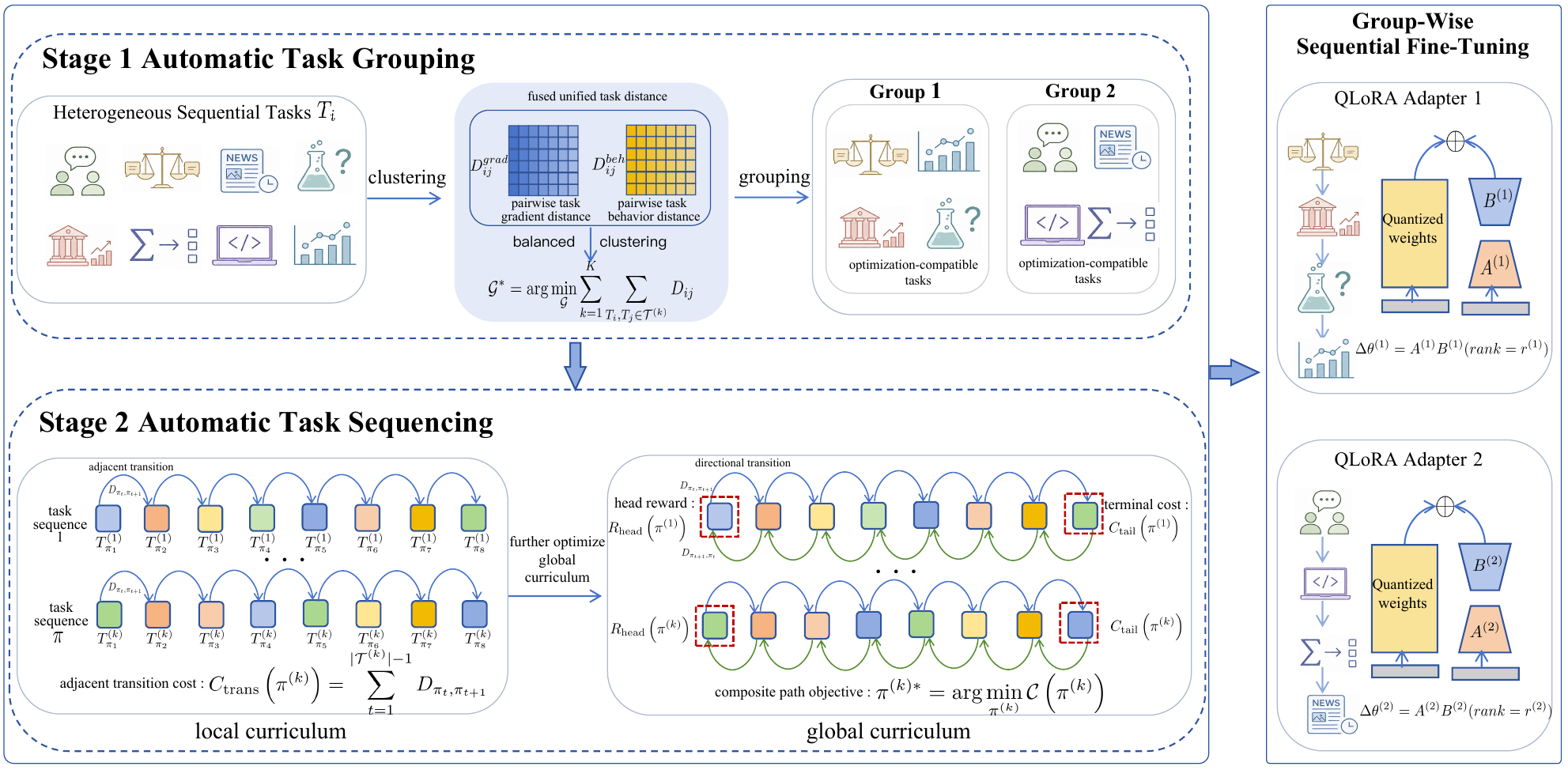}
    \caption{
    Our approach automatically organizes optimization paths through task grouping and task sequencing. These organized paths are implemented using independent QLoRA policies, reducing interference while preserving positive transfer. This indeed provides a new design
perspective for parameter-efficient fine-tuning over heterogeneous task sequences.}
    %{Overview of automated task sequencing for fine-tuning of Multi-policy LLMs. Heterogeneous tasks may follow incompatible optimization directions, causing severe interference in a shared adaptation space. Our approach automatically organizes optimization-compatible policies through task grouping and curriculum construction, reducing interference while preserving positive transfer under a fixed parameter budget.}
    \label{fig1}
\end{figure*}

\subsection{Automatic Task Grouping and Sequencing}
We propose an optimization-path organization framework for parameter-efficient fine-tuning over heterogeneous task sequences. The framework is implemented through a multi-policy PEFT architecture, where optimization-compatible task groups and task sequences are automatically constructed.
%We propose an automatic multi-policy PEFT framework that organizes heterogeneous continual adaptation under a fixed trainable parameter budget. 
Formally, the task collection is partitioned into $K$ disjoint groups:

\begin{equation}
\mathcal{T}
=
\bigcup_{k=1}^{K}\mathcal{T}^{(k)},
\qquad
\mathcal{T}^{(i)} \cap \mathcal{T}^{(j)}
=
\emptyset,
\quad i \neq j.
\end{equation}

For each task group $\mathcal{T}^{(k)}$, each organized optimization path is instantiated as an independent QLoRA adapter:
\begin{equation}
\Delta\theta^{(k)}
=
A^{(k)}B^{(k)},
\quad
A^{(k)} \in \mathbb{R}^{d\times r_k},
\quad
B^{(k)} \in \mathbb{R}^{r_k\times d}.
\end{equation}
Given an input $x$ from a task assigned to $\mathcal{T}^{(k)}$, the corresponding policy is activated:
\begin{equation}
f(x)
=
f_{\theta+\Delta\theta^{(k)}}(x).
\end{equation}

In QLoRA, there are two components: a frozen backbone and a set of trainable parameters that adapt the backbone based on the fine-tuning. In the multi-policy approach, each of the policies is initialized from the same frozen backbone, whereas the trainable adapter parameters are developed separately for each policy using the tasks grouped for that policy. %This design prevents strongly incompatible tasks from writing updates into the same low-rank space. At the same time, tasks assigned to the same policy continue to share an adapter and can therefore benefit from positive transfer.A trivial alternative would be to assign one independent adapter to every task. Although full task isolation can eliminate cross-task interference, it also removes the opportunity for positive transfer and causes the parameter cost to grow with the number of tasks.  
As shown in Figure \ref{fig1}, our process for developing fine-tuned policies happens in two steps: \textbf{Stage 1} determines how optimization paths are allocated by constructing optimization-compatible task groups. \textbf{Stage 2} determines how each optimization path evolves through  task sequencing. 
%In this sense, Stage 1 organizes \emph{where} tasks are optimized, whereas Stage 2 organizes \emph{how} optimization proceeds within each policy.

\subsubsection{Stage 1: Automatic Task Grouping}
%Unlike in a single policy approach, multiple policies allow for interfering tasks to be learnt separately. To do this practically requires a method to automate the grouping of these tasks to the different policies. 
This procedure attends to allocate optimization paths by grouping optimization-compatible tasks into independent adaptation spaces. These adaptation spaces are implemented as separate policies.
%The objective is to construct task groups with low internal optimization conflict, similar capability requirements, and compatible output behaviors.
The objective is to construct task groups with low internal optimization conflict, similar capability requirements, and compatible output behaviours, so that tasks within each policy are more likely to benefit from positive transfer while minimizing optimization interference.
These groups preserve positive transfer while separating strongly incompatible tasks into different adaptation spaces.

We first characterize each task according to its induced optimization behaviour. 
For each task $T_i$, gradients are collected from a temporary shared model over multiple mini-batches. Since the raw gradient space is extremely high-dimensional, principal component analysis (PCA) is applied to obtain a compact task representation $z_i^{\mathrm{grad}}$ while preserving the dominant variation among task-specific update directions. This parameter-free projection enables efficient pairwise task comparison. The pairwise gradient distance is then defined as,
\begin{equation}
D_{ij}^{\mathrm{grad}}
=
d
\left(
z_i^{\mathrm{grad}},
z_j^{\mathrm{grad}}
\right),
\end{equation}
where $d(\cdot,\cdot)$ denotes the normalized distance between task representations. 
%A small value indicates similar optimization tendencies, whereas a large value indicates potentially conflicting update directions. 
This distance directly estimates whether two tasks can safely co-update the same low-rank adaptation space. Note that the above gradient compatibility alone is insufficient for grouping heterogeneous language tasks. Two tasks may exhibit similar local gradient directions while requiring substantially different capabilities or output behaviors. Conversely, surface-level task similarity does not necessarily imply compatible optimization dynamics. Therefore a complementary behavior representation is constructed for each task. Specifically, dataset-level statistics are extracted to form a normalized behavior feature vector related to capability requirements and output structure, including prompt length, answer length, answer-to-prompt ratio, numeric-output tendency, multiple-choice tendency, short-answer tendency, reasoning-oriented prompts, and generation-oriented prompts. Let $z_i^{\mathrm{beh}}$ denote the resulting behavior representation. The corresponding pairwise behavior distance is,
\begin{equation}
D_{ij}^{\mathrm{beh}}
=
d
\left(
z_i^{\mathrm{beh}},
z_j^{\mathrm{beh}}
\right).
\end{equation}

The above distance captures structural differences that may not be fully reflected by instantaneous gradients, particularly differences in required capability and output format. The two compatibility signals provide complementary information. Gradient distance directly captures optimization conflict, whereas behavior distance captures differences in capability requirements and output structures. We therefore fuse them into a unified task distance:
\begin{equation}
D_{ij}
=
(1-\lambda)
D_{ij}^{\mathrm{grad}}
+
\lambda
D_{ij}^{\mathrm{beh}},
\end{equation}
where $\lambda \in [0,1]$ controls the contribution of behavior-level compatibility.
%This fusion is important because neither signal alone fully characterizes whether two heterogeneous tasks should share an adapter. Gradient similarity without behavioral compatibility may place structurally different tasks into the same policy, whereas behavioral similarity without gradient information may overlook direct optimization conflicts. Their combination provides a more comprehensive estimate of optimization compatibility.
Based on the fused distance matrix $D$, we construct $K$ task groups through balanced clustering:
\begin{equation}
\mathcal{G}^{*}
=
\arg\min_{\mathcal{G}}
\sum_{k=1}^{K}
\sum_{T_i,T_j\in\mathcal{T}^{(k)}}
D_{ij}.
\end{equation}

%In our setting, the balance constraint prevents degenerate solutions in which most tasks are assigned to one policy while only a small number of tasks are isolated in another. 
In the two-policy setting used in this work, we impose a maximum group-size constraint, resulting in two four-task groups on TRACE. This prevents one policy from absorbing most tasks while another serves only a small subset, thereby maintaining a more comparable task load under the fixed per-policy capacity. The constraint could be relaxed for larger task collections.
Consequently, Stage 1 constructs optimization-compatible adaptation paths whose internal tasks exhibit relatively low gradient conflict and similar capability requirements. Strongly incompatible tasks are separated into different adaptation spaces, while compatible tasks remain together to preserve positive transfer. 

\subsubsection{Stage 2: Automatic Task Sequencing}
%Task grouping alone does not fully solve heterogeneous continual adaptation. Although Stage 1 removes major cross-group conflicts, tasks assigned to the same adapter still update its parameters sequentially. Consequently, an unfavorable sequence can introduce destructive task transitions, overwrite early knowledge, or place a vulnerable task at an unsafe terminal position.  
Stage 2 further organizes how each optimization path evolves during sequential fine-tuning within each fixed group. Wang \citep{wang2023trace} demonstrate that optimizing the order in which tasks are presented can improve performance. However, their approach relies on manual search, which is impractical for fine-tuning because multiple permutations must be evaluated to identify an effective trajectory. We therefore develop an automated approach for constructing the task sequence.
For a task group $\mathcal{T}^{(k)}$, we search for a task sequence,
\begin{equation}
\pi^{(k)}
=
\left(
T_{\pi_1}^{(k)},
T_{\pi_2}^{(k)},
\ldots,
T_{\pi_{|\mathcal{T}^{(k)}|}}^{(k)}
\right),
\end{equation}
where $\pi^{(k)}$ defines the fine-tuning path followed by the $k$-th policy.
%\paragraph{Transition compatibility.}
A basic requirement for an effective task sequence is to avoid abrupt transitions between highly incompatible neighboring tasks. We therefore define the adjacent transition cost as,
\begin{equation}
C_{\mathrm{trans}}
\left(
\pi^{(k)}
\right)
=
\sum_{t=1}^{|\mathcal{T}^{(k)}|-1}
D_{\pi_t,\pi_{t+1}}.
\end{equation}
Minimizing this term ensures that neighboring tasks have compatible optimization characteristics, thereby reducing the immediate cost of moving from one task to the next. 

However, minimizing adjacent distance alone is insufficient. Sequential fine-tuning is directional, the effect of training $T_i$ followed by $T_j$ is not necessarily equivalent to training $T_j$ followed by $T_i$. Moreover, the first and final positions have distinct roles in the learning trajectory. 
A robust task sequence should consider not only local similarity but also the order in which tasks are learned.

%\paragraph{Head-task protection.}
Specifically, for the first task analysis, a head reward $R_{\mathrm{head}}\left(\pi^{(k)}\right)$ is introduced to encourage more complex tasks to be learned earlier, allowing the optimization path to establish complex capabilities before subsequent fine-tuning and helping reduce overall forgetting. Conversely, for the final task analysis, a terminal cost $C_{\mathrm{tail}}\left(\pi^{(k)}\right)$ is introduced to discourage selecting a task that may cause strong interference with previously learned tasks. This improves the stability of the final policy state.
 Directional transition effects are modelled through an interference cost $C_{\mathrm{dir}}\left(\pi^{(k)}\right)$ and a local transfer reward $R_{\mathrm{transfer}}\left(\pi^{(k)}\right)$, encouraging consecutive tasks to exhibit low optimization conflict while maintaining positive knowledge transfer. 
 However, a sequence composed only of locally compatible transitions may still form a poor global optimization path. We therefore further encourage smooth progression in task capability requirements $C_{\mathrm{prog}}\left(\pi^{(k)}\right)$.
This term discourages abrupt global changes in task complexity and promotes a coherent capability progression across the complete fine-tuning path. 
Combining these objectives, the final curriculum is obtained by minimizing the following composite path objective:
\begin{equation}
\begin{aligned}
\pi^{(k)*}
=
\arg\min_{\pi^{(k)}}
\mathcal{C}
\left(
\pi^{(k)}
\right),
\end{aligned}
\end{equation}
where
\begin{equation}
\begin{aligned}
\mathcal{C}
\left(
\pi^{(k)}
\right)
=&\;
C_{\mathrm{trans}}
+
 C_{\mathrm{tail}}
+
 C_{\mathrm{dir}}
+
 C_{\mathrm{prog}}
 \\
&-
 R_{\mathrm{head}}
-
 R_{\mathrm{transfer}}.
\end{aligned}
\end{equation}

%All objective terms are normalized and combined using fixed implementation weights, which remain unchanged across all experiments.The resulting curriculum jointly considers five aspects of continual adaptation: low transition cost, head-task protection, tail-task safety, local positive transfer, and global capability progression. All sequencing signals are derived automatically from task compatibility and behaviour characteristics, without relying on manually specified task sequences.

The objective terms are combined using fixed weights. The resulting optimization path jointly considers five aspects of sequential fine-tuning: low transition cost, head-task protection, tail-task safety, local positive transfer, and global capability progression. All sequencing signals are computed automatically from the optimization conflict matrix and behavioural statistics obtained during grouping, and are subsequently used to score candidate task sequences.

\subsection{Group-Wise Training}

After automatic grouping and sequencing, each policy is trained independently using its own QLoRA adapter. For group $\mathcal{T}^{(k)}$ with the selected sequence $\pi^{(k)*}$, the corresponding adapter is sequentially optimized along the task sequence:
\begin{equation}
T_{\pi_1}^{(k)}
\rightarrow
T_{\pi_2}^{(k)}
\rightarrow
\cdots
\rightarrow
T_{\pi_{|\mathcal{T}^{(k)}|}}^{(k)}.
\end{equation}

The group-wise objective can be written as,
\begin{equation}
\mathcal{L}_{\mathrm{multi}}
=
\sum_{k=1}^{K}
\sum_{T_i\in\mathcal{T}^{(k)}}
\mathbb{E}_{(x,y)\sim\mathcal{D}_i}
\left[
\ell_i
\left(
f_{\theta+\Delta\theta^{(k)}}(x),y
\right)
\right].
\end{equation}

To ensure that the advantage of the proposed framework does not simply result from using more trainable parameters, we explicitly control the total trainable capacity. Let $r$ denote the rank of a single shared-adapter baseline. The multi-policy configuration satisfies $\sum_{k=1}^{K} r_k = r$.
Consequently, any performance improvement under matched trainable capacity reflects the benefit of optimization-path organization rather than simple parameter expansion.

\section{Fine-Tuning Open-Source LLMs on TRACE dataset}

%\begin{table}[htb]
%\centering
%\small
%\caption{Controlled QLoRASeqFT settings. Cases B and C are matched in total trainable capacity.}
%\label{tab:settings}
%\begin{tabular}{lcccc}
%\toprule
%Case & Policy & Adapters & Rank & Tasks \\
%\midrule
%Case A & Single-policy & 1 & 64 & all 8 \\
%Case B & Multi-policy & 2 & 64 + 64 & 4 + 4 \\
%Case C & Single-policy & 1 & 128 & all 8 \\
%\bottomrule
%\end{tabular}
%\end{table}

\subsection{Benchmark and Experimental Setup}

We evaluate the proposed method on TRACE, a continual-learning benchmark for large language models \citep{wang2023trace} due to the wide range of different tasks for fine-tuning. TRACE contains eight heterogeneous tasks: C-STANCE, FOMC, MeetingBank, Py150, ScienceQA, NumGLUE-cm, NumGLUE-ds, and 20Minuten. 
%These datasets cover substantially different domains. 
While TRACE adopts LoRA, we use QLoRA through all experiments for its improved memory efficiency.
Experiments are conducted on the same two 7B-scale aligned chat models, namely LLaMA-2-7B-Chat and Vicuna-7B-V1.5. 
Although Vicuna is built upon LLaMA-2, it is instruction-tuned differently, providing an additional backbone to evaluate the robustness of our work.

\begin{table*}[htb]
\centering
\small
\caption{Comparison of different optimization-path organization strategies under the same trainable capacity on the TRACE benchmark using the LLaMA-2-7B-Chat and Vicuna-7B-V1.5 backbones. O-LoRA is included as a representative parameter-isolation multi-adapter PEFT baseline. * denotes results reproduced in our experimental setting.}
\label{tab:llama7b_main}
\begin{adjustbox}{max width=\textwidth}
\begin{tabular}{llllllcc}
\toprule
Backbone & Method & Policy &Group Strategy &Sequence Strategy & Setting & OP $\uparrow$ & BWT $\uparrow$ \\
\midrule
%Vicuna-7B-V1.5 & QLoRASeqFT & Single-policy &Manual& 1 shared rank-128 QLoRA & {39.5} & -0.05 \\
Vicuna-7B-V1.5 & QLoRASeqFT & Multi-policy & Random & Random & 2 rank-64 QLoRA & {36.49} & {-0.106} \\
% groupAB_vicuna_random_group_random_sequence_test1-test5
Vicuna-7B-V1.5 & QLoRASeqFT & Multi-policy & Auto & Random & 2 rank-64 QLoRA & {38.44} & {-0.083} \\
% groupAB_vicuna-auto-manual 13-17sh
{Vicuna-7B-V1.5} & QLoRASeqFT & Single-policy & N/A &Manual& 1 shared rank-128 QLoRA & {39.50} & -0.050 \\
Vicuna-7B-V1.5 & QLoRASeqFT & Multi-policy &Manual &Manual & 2 rank-64 QLoRA & {39.45} & \bf{}{-0.040} \\
% groupAB_vicuna_manual_v1-v5
Vicuna-7B-V1.5 & QLoRASeqFT & Multi-policy &Auto &Auto & 2 rank-64 QLoRA & \bf{41.14} & \bf{-0.040} 
\\
%groupAB_vicuna-auzre

\midrule

%average 5 runs, singleAdapter_baseline_v1-v5
LLaMA-2-7B-Chat & O-LoRASeqFT* &- &- &- &- &30.76 &-0.023 \\
LLaMA-2-7B-Chat & QLoRASeqFT & Multi-policy &Random &Random & 2 rank-64 QLoRA & {36.94 } & {-0.092} \\
LLaMA-2-7B-Chat & QLoRASeqFT & Multi-policy &Auto &Random & 2 rank-64 QLoRA & {40.75} & {-0.027} \\
{LLaMA-2-7B-Chat} & QLoRASeqFT & Single-policy & N/A &Manual& 1 shared rank-128 QLoRA & {42.12} & -0.041 \\
LLaMA-2-7B-Chat & QLoRASeqFT & Multi-policy &Manual &Manual & 2 rank-64 QLoRA & {44.53} & {0.012} \\
LLaMA-2-7B-Chat & QLoRASeqFT & Multi-policy &Auto &Auto & 2 rank-64 QLoRA & \bf{44.78} & \bf{0.013} \\
%average 5 runs, test 32-36
\bottomrule
\end{tabular}
\end{adjustbox}
\end{table*}

For each backbone, the main multi-policy setting contains two independent rank-64 QLoRA adapters. The total effective rank is 128. Different grouping and sequencing strategies use the same two-adapter architecture and trainable budget, allowing the effects of task grouping and task sequencing to be evaluated without changing the model capacity. 
We report Overall Performance (OP), i.e., the average performance over all tasks after fine tuning, and Backward Transfer (BWT) used to measure the average effect of later task training on previously learned tasks. Following the experimental setting of TRACE, each task uses 5,000 training samples. The remaining training hyperparameters are selected for our work: all tasks are trained for 3 epochs except for C-STANCE and NumGLUE-ds which are trained for 5 epochs, using a batch size of 2 and a learning rate of $1\times10^{-5}$.
%We also include the corresponding TRACE baselines for comparison under the same benchmark protocol. The training is kept the same as in \cite{wang2023trace}, where each dataset contains 5,000 training samples. Depending on the task, the training epoch is set to 3 or 5, the batch size is 2, and the learning rate is $1\times10^{-5}$. AdamW is used as the optimizer, and all experiments are conducted on a single NVIDIA H100 GPU.

%The validation and test sets follow the official data split and therefore vary in size across datasets.
%In Case B, the framework automatically constructs two optimization-compatible policies. The first adapter covers C-STANCE, FOMC, ScienceQA, and NumGLUE-ds, while the second covers MeetingBank, Py150, NumGLUE-cm, and 20Minuten. Stage 1 determines which tasks should share the same trainable subspace, and Stage 2 determines the continual learning sequence within each fixed group. Case B is therefore designed to test our main hypothesis: under a fixed parameter budget, organizing heterogeneous optimization paths can be more effective than forcing all tasks into one shared low-rank space.

\subsection{Comparison with Baseline Methods}
%For each backbone, we compare the proposed method with the TRACE baselines and the three controlled QLoRA settings. The comparison between Case A and LoRASeqFT evaluates the stronger QLoRA-based shared-adapter reference. Case B versus Case A tests the benefit of multi-policy organization, while Case B versus Case C distinguishes optimization-path organization from adapter scaling.
%O-LoRA \cite{wang2023orthogonal} is included as a representative external continual PEFT baseline, since it also addresses sequential adaptation by allocating separate low-rank subspaces to newly arriving tasks. Although O-LoRA is originally formulated for continual learning, it provides a relevant capacity-matched comparison for evaluating how low-rank adaptation spaces are organized across sequential tasks.This comparison is important because the remaining settings mainly investigate different designs within our multi-policy framework. 
O-LoRA \cite{wang2023orthogonal} is included as a parameter-isolation multi-adapter PEFT baseline, assigning one independent LoRA subspace to each individual task, whereas our framework automatically groups optimization-compatible tasks into shared policy-specific adaptation spaces. This makes O-LoRA a particularly appropriate baseline for isolating the effect of different optimization-path organization strategies.
To ensure a fair comparison under the same total low-rank budget, O-LoRA allocates a rank-16 task-specific block to each of the eight tasks, resulting in a total rank of 128, while our framework uses two rank-64 policy-specific adapters, also resulting in a total rank of 128. Under this capacity-matched setting, O-LoRA achieves an OP of 30.76 and a BWT of -0.023 on LLaMA-2-7B-Chat, which is substantially lower than our automatic framework implemented with multi-policy PEFT. This indicates that, under the same total trainable capacity, organizing compatible tasks into shared policy-specific adaptation spaces is more effective than allocating an independent small subspace to each individual task.

%For completeness, we also evaluate a capacity-matched single-policy baseline using one shared rank-128 QLoRA adapter. On LLaMA-2-7B-Chat, this baseline achieves an OP of 37.69 with a BWT of -0.083, while on Vicuna-7B-V1.5 it achieves an OP of 39.50 with a BWT of -0.050. Compared with this baseline, the proposed automatic multi-policy framework achieves higher OP on both backbones under the same total trainable capacity, indicating that organizing optimization paths is more effective than simply increasing the capacity of a shared adapter.

As shown in Table~\ref{tab:llama7b_main}, we evaluate a capacity-matched single-policy baseline using 1 shared rank-128 QLoRA adapter. Compared with this baseline, the proposed automatic multi-policy framework achieves higher OP on both backbones under the same total trainable capacity, indicating that organizing optimization paths is more effective than simply increasing the capacity of a shared adapter. We further compare four grouping-and-sequencing settings under the same two-adapter architecture and trainable budget. Random grouping with random sequencing represents a multi-policy baseline without informed task organization. Automatic grouping with random sequencing isolates the effect of the proposed grouping strategy. The manually designed grouping and task sequence represent an expert-designed solution, where task assignment and learning sequence are selected using prior knowledge and manual analysis. The complete framework combines automatic grouping with automatic task sequencing, with the aim of achieving optimization-path organization comparable to or better than the expert-designed solution without requiring manual intervention. 

On LLaMA-2-7B-Chat, we can find that replacing random grouping with automatic grouping improves OP to 40.75 and BWT to -0.027, even though the task sequence remains random. This shows that automatically assigning optimization-compatible tasks to the same policy already provides a substantial improvement in both final performance and knowledge retention. Compared to the expert-designed grouping and sequence, the complete automatic framework obtains the best result, reaching an OP of 44.78 and a BWT of 0.013. This shows that the automatic framework reaches and slightly exceeds expert-level optimization-path organization without requiring prior knowledge of the task relationships or manual search over possible groupings and sequences. A similar overall trend is observed on Vicuna-7B-V1.5. Random grouping and random sequencing achieve an OP of 36.49 and a BWT of -0.106, while the expert-designed setting improves OP to 39.45 and BWT to -0.040. The complete automatic framework achieves the best OP of 41.14 while maintaining the same BWT of -0.040 as the expert-designed solution. 
Consequently, the advantage of the proposed framework does not come from introducing multiple adapters themselves, but from organizing optimization paths through automatic task grouping and  task sequencing.
%Consequently, the advantage of the proposed framework does not come only from introducing multiple adapters, but from automatically constructing effective task groups and learning trajectories.

\begin{figure}[h]
    \centering
    \includegraphics[width=0.9\linewidth]{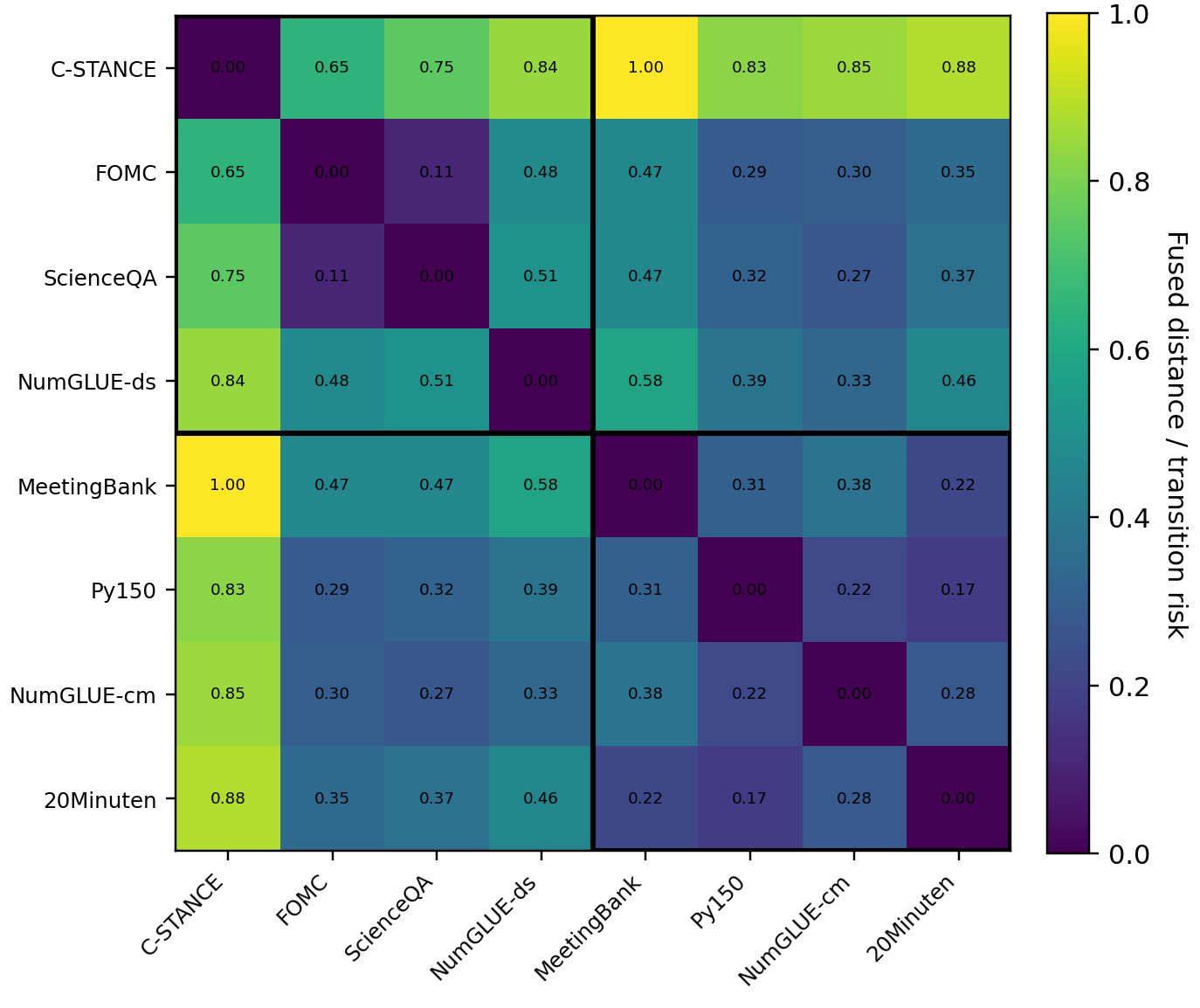}
    \caption{Visualization of the fused task-compatibility structure used for automatic grouping. Lower pairwise distances indicate more compatible optimization directions and behavior characteristics, providing the basis for constructing the two optimization-compatible policies.}
    \label{fig2}
\end{figure}

\begin{table*}[t]
\centering
\caption{
Task-level analysis of destructive interference under different task grouping and sequencing strategies. Lower BWT indicates stronger destructive interference. ``--'' denotes the last task within a policy, for which BWT is undefined.
}
\label{tab:task_level_bwt}
\resizebox{\textwidth}{!}{
\begin{tabular}{lcccccccc}
\toprule
\textbf{Grouping / sequencing}
& \textbf{C-STANCE}
& \textbf{FOMC}
& \textbf{ScienceQA}
& \textbf{NumGLUE-ds}
& \textbf{MeetingBank}
& \textbf{Py150}
& \textbf{NumGLUE-cm}
& \textbf{20Minuten}\\
\midrule

Random / Random
& -0.002 & -- & -0.046 & -0.030 & \textbf{-0.357} & -0.018 & \textbf{-0.256} & --
\\

Auto / Random
& 0.000 & 0.109 & 0.052 & -0.012 & \textbf{-0.048} & -- & \textbf{-0.231} & -0.004
\\

Auto / Auto
& 0.009 & 0.107 & 0.135 & -- & \textbf{-0.171} & -0.014 & \textbf{0.000} & --
\\

\bottomrule
\end{tabular}
}
\end{table*}

\begin{table*}[htb]
\centering
\small
\caption{Ablation on task grouping strategy. 
``Auto Group'' denotes the task-compatibility-aware grouping used in the main
experiments, which is compared with two alternative groupings derived from
the fused task-compatibility structure.}
\label{tab:grouping_ablation}

\setlength{\tabcolsep}{3.5pt}
\renewcommand{\arraystretch}{1.08}

\begin{adjustbox}{max width=\textwidth}
\begin{tabular}{@{}lllcc@{}}
\toprule
Grouping
& Group 1
& Group 2
& OP $\uparrow$
& BWT $\uparrow$ \\
\midrule

Alternative 1
& \{C-STANCE, FOMC, NumGLUE-ds, \textbf{20Minuten}\}
& \{MeetingBank, Py150, NumGLUE-cm, \textbf{ScienceQA}\}
& 41.29
& -0.017 \\

Alternative 2
& \{C-STANCE, FOMC, NumGLUE-ds, \textbf{MeetingBank}\}
& \{Py150, \textbf{ScienceQA}, NumGLUE-cm, 20Minuten\}
& 41.96
& -0.003 \\

Auto Group
& \{C-STANCE, FOMC, ScienceQA, NumGLUE-ds\}
& \{MeetingBank, Py150, NumGLUE-cm, 20Minuten\}
& \textbf{44.78}
& \textbf{0.013} \\

\bottomrule
\end{tabular}
\end{adjustbox}
\end{table*}

\subsection{Analysis of Optimization-Path Organization}
In this part, we analyze three components of optimization-path organization: task grouping, rank allocation, and task sequencing.

\textbf{Task grouping.}  
To evaluate the proposed grouping strategy, we compare it with two alternative task groupings. All settings use the same two-adapter architecture, the same rank allocation, and the same sequential fine-tuning protocol. Only the task assignments are changed. 
As shown in Table 3, compared to the two alternative groupings, the proposed grouping achieves the best OP of 44.78 and the best BWT of 0.013. This indicates that transfer learning has been effective, with methods such as MTL-LoRA \cite{yang2025mtlloralowrankadaptationmultitask} and MoRE \cite{zhang-etal-2025-mixture} having a maximum BWT of 0. Since all other settings remain unchanged, the improvement comes from organizing optimization-compatible tasks into the same policy rather than simply introducing multiple adapters. Moreover, Figure 2 is used to visualize the fused task-compatibility matrix used in Stage 1, which combines gradient conflict and behavior differences to organize optimization-compatible tasks into two policies.
Figure 2 and Table 3 demonstrate the necessity of task grouping for effective parameter-efficient fine-tuning over heterogeneous task sequences.

\textbf{Rank allocation.} We further analyze how the fixed trainable budget should be allocated across the two policies. The total effective rank is fixed at 128, while only the rank allocation between the two adapters is changed. We compare three settings: 32 + 96, 64 + 64, and 96 + 32. As shown in Table 4, the balanced 64 + 64 allocation achieves the best OP and BWT. This indicates that the performance improvement does not depend on allocating more trainable parameters to one policy. Instead, balancing the trainable capacity across the two policies provides a more effective use of the fixed PEFT budget.

\textbf{Task sequencing.} The task grouping, rank allocation, and training configuration remain unchanged, while only the task sequence within each policy is changed. As shown in Table 5, the automatic sequence achieves the highest OP and BWT for both groups, while both random sequences reduce OP and further increase forgetting. These results indicate that task grouping alone is insufficient. Organizing the optimization trajectory within each policy further improves sequential fine-tuning performance and reduces catastrophic forgetting.

Overall, these results demonstrate that the effectiveness of the proposed framework depends on the above components. Meaningful task grouping organizes optimization-compatible tasks into the same adaptation space, balanced rank allocation makes effective use of the fixed trainable budget, and automatic task sequencing further improves fine-tuning performance and knowledge retention.

\begin{table}[htb]
\centering
\small
\caption{Ablation on rank allocation across policies under a fixed total trainable budget.The total LoRA rank is fixed at 128 for all settings.}
\label{tab:scaling_vs_decoupling}
\begin{adjustbox}{max width=\columnwidth}
\begin{tabular}{lccc}
\toprule
Rank allocation (Group 1 + Group 2) & Total rank & OP $\uparrow$ & BWT $\uparrow$ \\
\midrule
{32 + 96} & 128 & 41.61 & -0.018 \\
{96 + 32} & 128 & 42.58 & -0.011 \\
{64 + 64} & 128 & \textbf{44.78} & \textbf{0.013} \\
\bottomrule
\end{tabular}
\end{adjustbox}
\end{table}

\begin{table}[htb]
\centering
\small
\caption{Ablation on task sequencing strategy. Task grouping is fixed using the optimization-compatible groups generated by our method.}
\label{tab:sequence_sensitivity}
\begin{adjustbox}{max width=\columnwidth}
\begin{tabular}{lcccc}
\toprule
sequence & Group1-OP & Group1-BWT & Group2-OP & Group2-BWT \\
\midrule
{Random sequence 1} & 54.86 & 0.054 & {25.89} & {-0.119} \\
{Random sequence 2} & 54.02 & 0.049 & {27.75} & {-0.094} \\
{Auto sequence} & \textbf{55.81} & \textbf{0.084} & \textbf{33.90} & \textbf{-0.062} \\
\bottomrule
\end{tabular}
\end{adjustbox}
\end{table}

\subsection{Analysis of Destructive Interference}
Table~\ref{tab:task_level_bwt} provides a task-level analysis of destructive interference under different optimization-path organization strategies. Under random grouping and sequencing, MeetingBank and NumGLUE-cm show the largest negative backward transfer, indicating severe interference from later tasks. Automatic grouping greatly reduces interference on MeetingBank, while automatic sequencing further improves ScienceQA and removes forgetting on NumGLUE-cm. Overall, our method does not eliminate interference for every task, but improves the overall distribution of transfer effects across tasks, leading to stronger positive transfer among compatible tasks.

\section{Conclusion}

%In this paper, we propose an automatic task sequencing multi-policy PEFT framework for heterogeneous continual adaptation of large language models. 
In this paper, we proposed an optimization-path organization framework for parameter-efficient fine-tuning of large language models, implemented using a multi-policy PEFT architecture.
Instead of forcing heterogeneous tasks to share a single low-rank adaptation space, our framework automatically organizes multiple optimization-compatible adaptation paths through task grouping and task sequencing. 
%Specifically, task grouping determines which tasks should share an adapter by reducing within-policy gradient conflict, capability differences, and output-format mismatch, while group-internal sequencing constructs a continual learning path that reduces transition costs and promotes stable knowledge transfer and capability progression. 
The approach separates incompatible tasks into independent adaptation spaces, reducing interference while preserving positive transfer among compatible tasks. This provides a new design perspective for parameter-efficient fine-tuning over heterogeneous task sequences, where optimization-path organization can be more effective than simply increasing adapter capacity.

\medskip

\bibliography{bib}

\newpage

\end{document}